\pdfoutput=1

\documentclass[11pt]{article}

\usepackage{acl2023}

\usepackage{times}
\usepackage{latexsym}

\usepackage[T1]{fontenc}

\usepackage[utf8]{inputenc}

\usepackage{microtype}

\usepackage{inconsolata}

\usepackage{longtable}
\usepackage{tabularx}
\usepackage{array}
\usepackage{bm}
\usepackage{graphicx}
\usepackage{soul}
\usepackage {arydshln}
\usepackage{amsmath}

\usepackage{subcaption}

\title{SumRec: A Framework for Recommendation using Open-Domain Dialogue}

\author{Ryutaro Asahara, Masaki Takahashi, Chiho Iwahashi, Michimasa Inaba \\
        Inaba Lab, Department of Informatics, The University of Electro-Communications \\
        \texttt{r-asahara@uec.ac.jp, \{t171m2.dt, i2130008\}@mail.uec.jp, m-inaba@uec.ac.jp}}

\begin{document}
\maketitle
\begin{abstract}
Chat dialogues contain considerable useful information about a speaker’s interests, preferences, and experiences.
Thus, knowledge from open-domain chat dialogue can be used to personalize various systems and offer recommendations for advanced information.
This study proposed a novel framework SumRec for recommending information from open-domain chat dialogue.
The study also examined the framework using ChatRec, a newly constructed dataset for training and evaluation \footnote{Our dataset and code is publicly available at \url{https://github.com/Ryutaro-A/SumRec}}. 
To extract the speaker and item characteristics, the SumRec framework employs a large language model (LLM) to generate a summary of the speaker information from a dialogue and to recommend information about an item according to the type of user.
The speaker and item information are then input into a score estimation model, generating a recommendation score.
Experimental results show that the SumRec framework provides better recommendations than the baseline method of using dialogues and item descriptions in their original form.
\end{abstract}

\section{Introduction}

Chat dialogues, which contain substantial information about a speaker’s interests, preferences, and experiences, account for an estimated 62\% of human dialogue \cite{koiso2016survey}. 
Information from open-domain chat dialogues can be used to personalize and suggest advanced information in various systems; however, such a research direction has been negligibly explored. 
Inaba et al. proposed a method for estimating users’ interests through open-domain chat dialogues \cite{inaba2018estimating}, although they still need to address how to use the estimated interests. 
Hence, this study suggests a framework for recommending information from open-domain chat dialogues, focusing on tourist spots as the recommendation target. 
\begin{figure}[t]
\centering
\includegraphics[width=\linewidth]{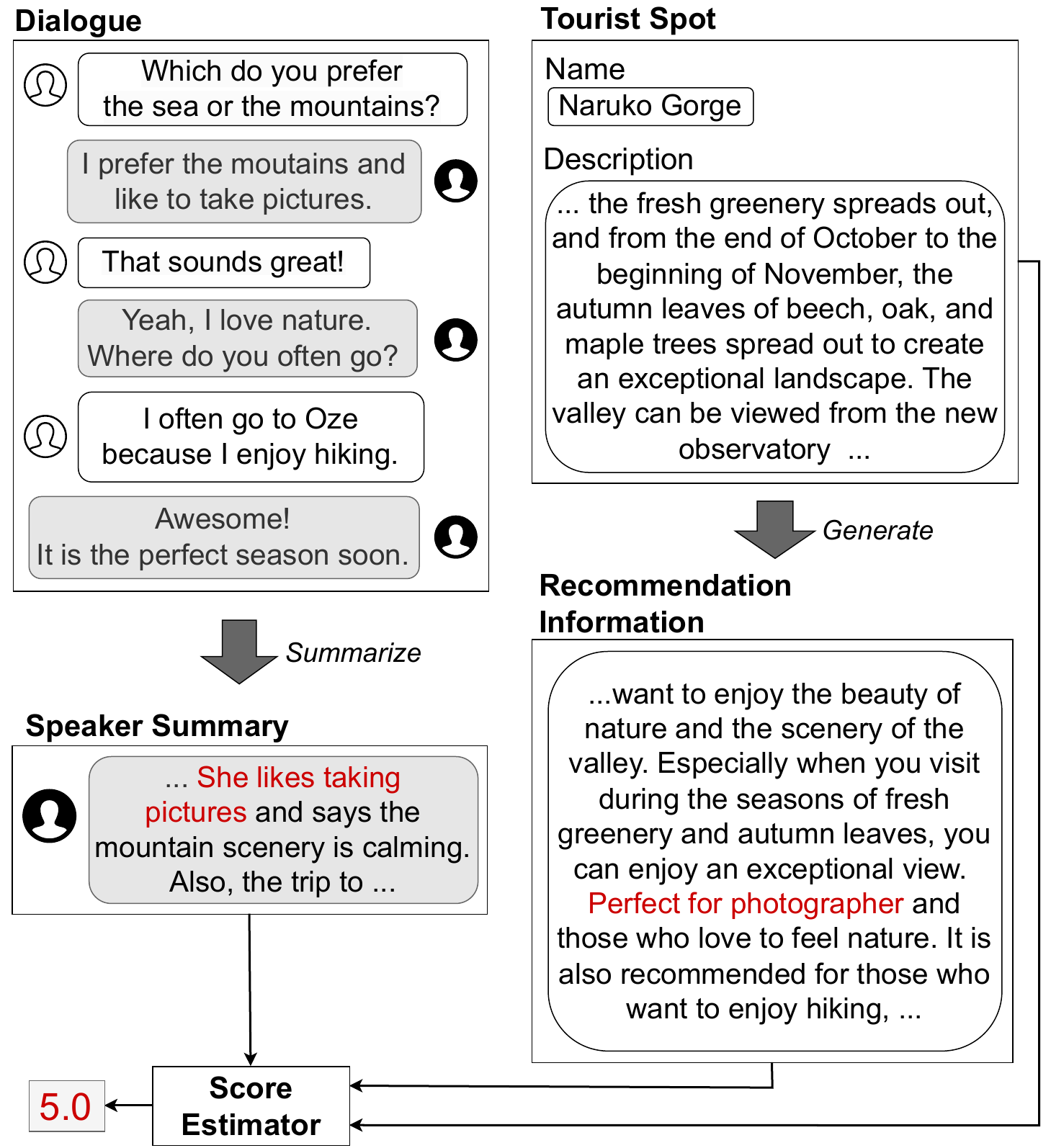}
\caption{
Our task and the SumRec framework: the task aims to predict the speaker’s score based on given chat dialogue and tourist-spot information.
The SumRec framework generates speaker information from the dialogue and recommendation information from the tourist-spot description and then estimates the score.
}
\label{fig:gaiyo}
\centering
\end{figure}
For example, a user who is interested in local cuisine can obtain restaurant recommendations, one who is interested in history can obtain recommendations for historical buildings, and one who is interested in movies can obtain recommendations for places where movies were filmed.
The challenge in such tasks lie in how the speaker’s information in the chat dialogue can be linked to information about tourist spots.
Specifically, in domains where external knowledge, such as knowledge graphs, is not available, relations between entities cannot be explicitly managed.
Therefore, we propose a framework that employs a large language model (LLM) to associate items with chat dialogues and offer recommendations without using external knowledge (Figure \ref{fig:gaiyo}).

We also construct a dataset ChatRec and use it to evaluate the proposed framework. Our contributions are as follows:
\begin{itemize}
      \item We create a novel recommendation task based on chat dialogues and propose the SumRec framework to manage it.
      \item To facilitate analysis of the recommendation task, we construct a dataset called ChatRec to recommend tourist spots based on chat dialogues and collect human recommendation data that are not available in other conversational recommender system (CRS) datasets.
      \item We propose SumRec for this task and demonstrate its effectiveness through evaluation experiments using ChatRec.
\end{itemize}

\section{Related Work}
\paragraph{Conversational Recommendation}
Our research is associated with studies on CRSs, whose goal is to determine users’ preferences and experiences while conversing with them and offer recommendations based on such conversations. Several datasets have been released to develop CRSs.

For example, I{\small NSPIRED} dataset \cite{hayati2020inspired} contains dialogues centered on film recommendations, with each dialogue involving two participants—a recommender and a seeker. 
The recommender’s task is to comprehend the seeker’s taste and suggest suitable movies. The model, which relies on previous research \cite{b00b5cbf6ca947d0ae7f1bd5566dd3b1,10.2753/jec1086-4415110204,dowrick199923,10.1145/764008.763957} with regard to human behavior–concerning recommendations, employs dialogue tactics developed to craft informed strategies for suggestions. 
Nonetheless, associating the speaker’s (i.e., seeker’s) preferences with a film’s traits and offering a suitable recommendation presents a notable challenge.

Next, DuRecDial \cite{liu2020towards} is a Chinese-language recommendation dialogue dataset comprising natural dialogues in which the recommender recommends as well as chats and answers questions. 
In the suggested model, a response is generated or chosen based on the conversation history and the information available.
Nevertheless, difficulties arise when no relevant items are found within the dialogue content and informative sentences. 
This can inhibit the ability to provide recommendations appropriately tailored to the speaker’s preference.

As shown above, the domain of dialogues and recommendation targets are the same in existing datasets.
In our task, the dialogue and recommendation target domains are not always identical. In the task involving recommendation dialogues, the model can identify the recommendation targets based on entities appearing in the dialogues, such as movie titles or director names.
Conversely, the dialogues in our task do not necessarily concern travel or tourist destinations.
The model must utilize information unrelated to tourist spots, such as the speaker’s hobbies and experiences, to provide recommendations.

A third example is FoCus \cite{jang2022customized}, a dialogue dataset that provides background information about the user, that is, “the persona,” alongside details about tourist spots.
The persona encapsulates details about users’ experiences, preferences, property, hobbies, and interests.
The objective is to enable the system to engage in suitable dialogues, considering the persona and the associated tourist attraction information.
The proposed model delves into the user persona and tourism knowledge for interactive dialogue.
However, the model operates on predefined personas and fails to account for possible application to implicitly characterized personas that emerge during the dialogue.

\paragraph{Speaker Information Extraction}
Studies have examined speaker information extraction from dialogue.
Some have extracted entities and relations between entities in dialogue \cite{bang2015example,hirano2015user}. Meanwhile, others have estimated users’ interests \cite{inaba2018estimating}, persona \cite{zhong-etal-2022-less}, and implicit profiles \cite{ma2021one}.
Unlike the present research, these studies sought to personalize dialogue systems and not extend to domains outside dialogue.

The proposed framework extracts speaker information through a dialogue summary. Dialogue summarization has been actively studied \cite{zhang2021unsupervised,feng-etal-2021-language,li-etal-2023-dionysus}, with several datasets being developed \cite{gliwa2019samsum,chen2021dialogsum,shin-etal-2022-dialogue}.
However, no studies have been conducted on using dialogue summaries for recommendations.

\paragraph{Explainable Recommendation}
The proposed framework generates a description of the type of users for which a tourist spot is recommended.
Studies have been conducted on the task of generating reasons for recommendations, such as obtaining similar reviews and aspects using a search module \cite{cheng-etal-2023-explainable} and imposing lexical constraints on the generated recommendation sentences \cite{li2023ucepic}.

However, these studies aimed to perform sentence generation using user behavior history and review data and did not consider the application of recommendation sentences using dialogue data.

\section{SumRec Framework}
\begin{figure}[t]
\begin{center}
\includegraphics[scale=0.5]{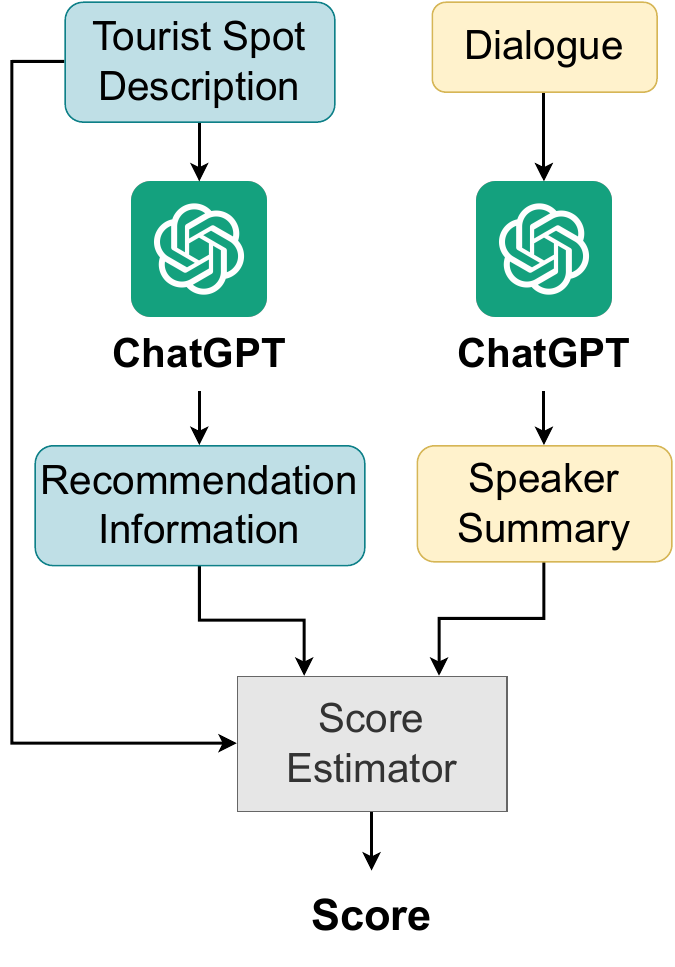}
\caption{Pipeline of SumRec framework}
\label{fig:frame_work}
\end{center}
\end{figure}

Chat dialogue typically includes valuable details, such as the speaker’s interests, preferences, experiences, and routines, which are critical for identifying a recommendation subject.
Meanwhile, much information, such as greetings and questions to the other party, is unnecessary for a recommendation.
The subject matter of a dialogue is not always tourism.
Such situations complicate the task of linking speaker information to the intended tourist destination recommendation.
Hence, we introduce a framework based on a straightforward concept.
We use an LLM to generate speaker summaries from chat dialogues and provide recommendation information based on tourist-spot descriptions.
We then propose the SumRec framework for rating tourist spots based on these descriptions.
Figure \ref{fig:frame_work} presents an overview of our proposed framework.
The speaker summary effectively recaps the speaker’s hobbies, preferences, experiences, and routine behavior.
The recommendation information describes the appropriate users for a tourist spot.
Using such data allows the model to effectively link speaker information with tourist-spot data, facilitating suitable suggestions.

\subsection{Task Definition}
Consider a text chat dialogue $U = \{u_1, u_2, ..., u_N\}$ between two workers A and B, tourist spots $V = \{v_1, v_2, ..., v_M\}$, and tourist spot descriptions $E = \{e_1, e_2, ..., e_M\}$.
$u_i$ is the $i$ th utterance in the dialogue, $N$ is the number of utterances in the dialogue, and $M$ is the number of tourist spots.
The goal of this task is to predict worker A’s evaluation score for the tourist spots.
$T^A = \{t^A_{v_1}, t^A_{v_2}, ..., t^A_{v_M}\}$, and worker B's evaluation scores $T^B = \{t^B_{v_1}, t^B_{v_2}, ..., t^B_{v_M}\}$ from the dialogue $U$, descriptions of the tourist spots $E$.

\subsection{Speaker Summary}
Dialogues contain much information that is not directly related to recommendations, such as greetings and backchanneling, which may hinder the delivery of appropriate recommendations.

Therefore, this study employs an LLM to generate speaker summaries containing information from chat dialogues, such as users’ hobbies, preferences, habits, and experiences, which are useful for recommendation.
Dialogue $U$ is inputted into the LLM to generate A’s summary $s_A$ and B’s summary $s_B$.
This study uses few-shot prompting in ChatGPT as its LLM.
ChatGPT generates $s_A$ and $s_B$ from a task description, an example of a dialogue and its summary, and the target dialogue U for summary generation.
Table \ref{summary_prompt} (Appendix \ref{Promt Design}) presents an example of a prompt and a generated summary.

\subsection{Recommendation Information}
The description of a tourist spot usually focuses on its features, with limited information about the types of visitors. 
Consequently, crucial data that can link a speaker’s summary to the description of such tourist spots are scarce, making recommendation a challenging task.

This study used an LLM to generate recommendation information $r_i(i=1,2,...,M)$, which is an extension of tourist spot description $e_i(i=1,2,...,M)$. 
The recommendation information $r_i$ includes information on what type of travelers it is suitable for tourist spot $i$.
This information facilitates an interaction between the proposed information and the preferences and habits in the speaker’s summary.
Therefore, this approach allows the model to deliver tailored recommendations.
We designed few-shot prompts to generate recommendation information. This means that ChatGPT generates $r_i$ from a task description, five tourist-spot descriptions, examples of their recommendation information, and a description $s_i$ as a target.
Table \ref{recommend_prompt} (Appendix \ref{Promt Design}) presents an example of prompts and generated information.

\subsection{Score Estimator}
We use speaker summary $s_k(k=A,B)$, tourist-spot description $e_i$, and recommendation information $r_i$ obtained through the above procedure as inputs for the score estimator, which then provides a predicted evaluation score $y^k_i$.

The score estimator is available for any language model regardless of its properties.
We conduct two experiments: fine-tuning a pretrained RoBERTa model and generating evaluation scores in ChatGPT via few-shot prompting.

A more effective model for fine-tuning architectures would be a subject of future work.

\section{Experiment}
\subsection{ChatRec Dataset}

The ChatRec dataset includes chat dialogues between two speakers as well as information on each evaluation of various tourist spots.

\subsubsection{Tourist Spots}

We use data from Rurubu\footnote{https://solution.jtbpublishing.co.jp/service/domestic/} and TripAdvisor to obtain information about tourist spots. 
Rurubu contains approximately 45,000 Japanese tourist spots, while TripAdvisor is a platform that features user reviews and price comparisons for travel-related services, such as hotels and restaurants.

The Rurubu data comprises professionally written descriptions, addresses, and operating hours of various tourist spots.
As Rurubu data include nationwide chain stores and restaurants as tourist spots, we excluded the inappropriate ones using reviews from TripAdvisor.
The tourist-spot selection process is as follows: (1) collect all tourist spots from Rurubu data and (2) exclude entries with <100 user reviews on TripAdvisor.
This procedure yields 3,290 tourist spots and their Descriptions, which are collected in the Rurubu data.
The tourist-spot descriptions have a mean word count of 96.91.

To evaluate the tourist spots, we create 147 files from these locations and descriptions. Each file contains data on 10–20 spots, and tourist spots in the same file are located in the same prefecture in Japan.

\subsubsection{Dialogue and Evaluation Score Collection} \label{score}
We collect two-person text chat dialogues using CrowdWorks\footnote{https://CrowdWorks.jp}—a crowdsourcing service. 
Each worker speaks in turn, and the dialogue is completed when each worker has spoken at least for 10 turns.
For the dialogue scenario, we inform the workers that they are talking to a stranger sitting next to them in a waiting room.

With regard the topic, as the workers know that they will be asked to evaluate tourist spots at the end of the dialogue, we expect that the topic would be biased toward travel if we provide no instructions.
Therefore, we set three parameters regarding the topics that the participants are allowed to discuss:
(1) focusing on a travel-related topic (\textbf{Travel}),
(2) centered on any topic except travel (\textbf{Except for Travel}), and 
(3) no limitations on the topic (\textbf{No Restriction}).

After the dialogue, 1 of the 147 files for tourist-spot evaluation (created as shown in the previous section) is randomly assigned to the workers, and each worker fills out the file with their evaluation score for the spots.
Both workers are given the same file and asked to provide their ratings based on a five-point scale from 1 (I don’t like there./I don’t want to go there.) to 5 (I like there./I want to go there.).

To evaluate the recommendation performance of humans for tourist spots, we collect scores from third-party workers.
Their instructions are as follows: (1) read a dialogue and all the descriptions of tourist spots, (2) predict each speaker’s interest level toward the given tourist spot, and (3) assign a score to each destination on a scale of 1–5.
Five workers are assigned to one dialogue, and we use the average of the five workers’ predicted scores in the experiment.

\begin{figure}[t]
\begin{center}
\includegraphics[width=\linewidth]{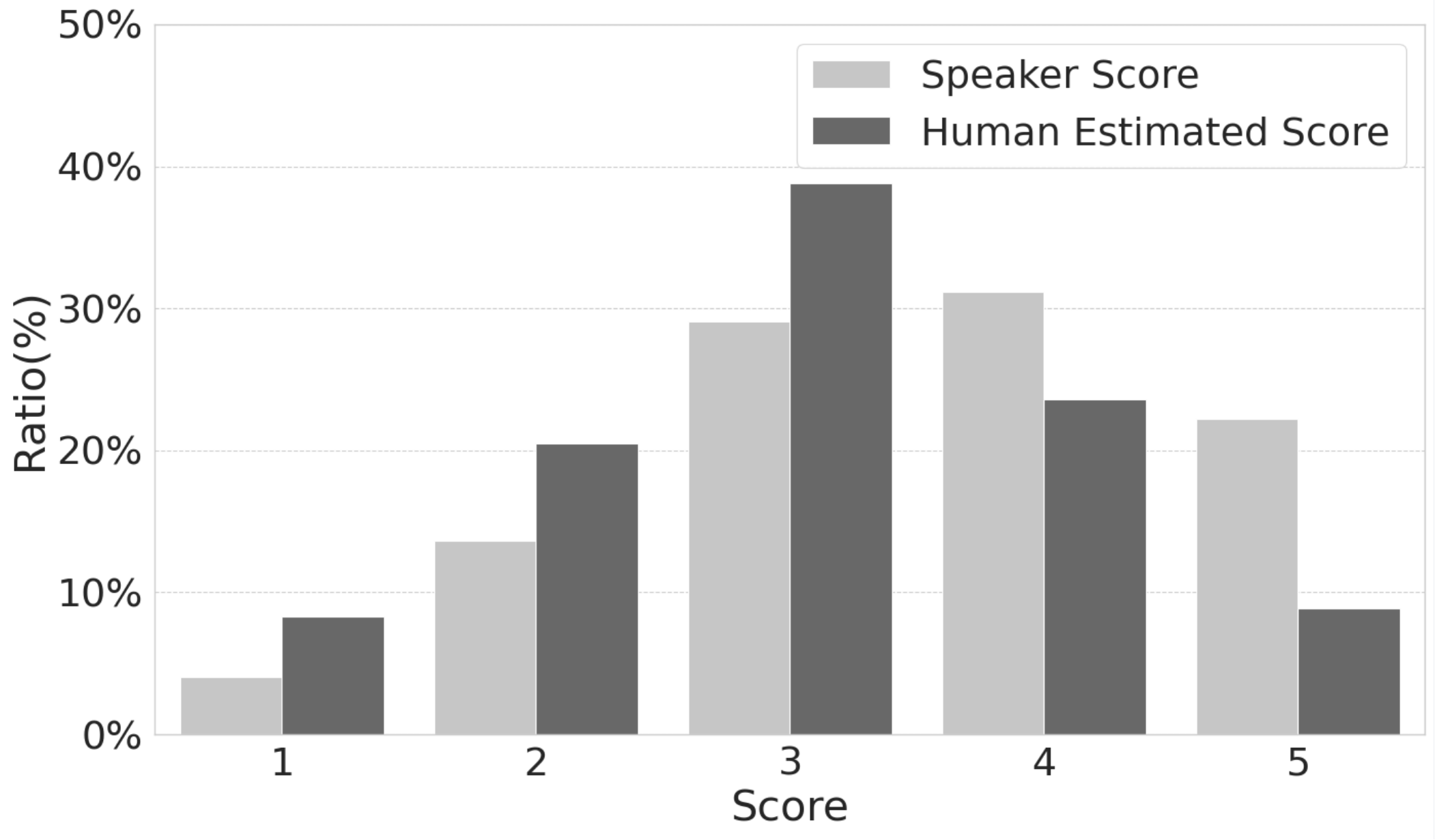}
\caption{
Distribution of speaker and human estimated scores in the ChatRec dataset.
}
\label{human-dst-graph}
\end{center}
\end{figure}

We obtain 1,005 dialogue datapoints and the associated tourist-spot evaluation data. 
Table \ref{data-statistics} presents the statistics of the data according to the following settings: ``Travel (T),'' ``Except for Travel (E),'' and ``No Restriction (N).''
Moreover, Figure \ref{human-dst-graph} presents the distribution of the speaker’s own evaluation and the third party’s prediction.
The estimated scores exhibit a balanced distribution around 3, whereas the speaker scores trend higher.

\begin{table}[t]
\centering
\begin{tabularx}{78mm}{p{24mm}p{8mm}p{8mm}p{8mm}p{8mm}}
\hline
\textbf{} & \textbf{T} & \textbf{E} & \textbf{N} & \textbf{ALL} \\
\hline
Dialogues & 237 & 223 & 545 & 1,005 \\
Utterances & 5,238 & 5,009 & 11,735 & 21,982 \\
Words per Uttr. & 21.56 & 23.96 & 15.77 & 23.44 \\
Spots per Dial. & 15.75 & 15.62 & 15.77 & 15.73 \\
\hline
\end{tabularx}
\caption{
Statistics of the ChatRec dataset. T, E, and N denote three types of dialogue topics: Travel, Except for Travel, and No Restriction, respectively.
}
\label{data-statistics}
\end{table}

\subsubsection{Dataset Splitting}
We implement an 8:1:1 split for train, validation, and test data.
Using the Rurubu data, we analyze the category distribution of each tourist spot, such as historical buildings, nature, and restaurants, and divide it into train, validation, and test so that the distribution is as balanced as possible.

\subsection{Implementation Details}
We used RoBERTa and ChatGPT as score estimators.

For the RoBERTa-based score estimator, we used the Japanese RoBERTa$ _{large}$\footnote {https://huggingface.co/nlp-waseda/RoBERTa-large-japanese-seq512-with-auto-jumanpp}.
We encoded speaker summaries and tourist-spot information (tourist-spot descriptions and recommendation information concatenated by the $[SEP]$ token) using the bi-encoder scheme. 
RoBERTa individually receives the speaker summaries and tourist-spot information with $[CLS]$ tokens at the beginning.
The linear layer receives each concatenated output vector corresponding to the $[CLS]$ tokens and predicts the score.
To find the optimal hyperparameters, we performed a grid search.
As a result, we set a learning rate of $1e^{-6} $, a batch size of 32, and a maximum of 10 epochs and conducted early stopping for 3 epochs using the Adafactor optimizer.
For the evaluation, we used the checkpoint with the lowest loss score on the validation set.

ChatGPT generated an estimated score $y^k_i$ from a task definition sentence, a five-shot sample, speaker summary $s_k(k=A,B)$, and tourist-spot description $e_i$ and recommendation information $r_i$. Table \ref{sumrec_prompt} (Appendix \ref{Promt Design}) presents an example of the prompt.

\subsection{Comparison Method}
We compared the performance of our proposed framework with those of four baseline methods.

\paragraph{Random}
This randomly determines tourist spot scores without dialogue and tourist spot descriptions.

\paragraph{RoBERTa}
This method involved feeding a dialogue $U$ and a description of the tourist spot $e_i$ directly into RoBERTa. As a dialogue $U$ includes utterances from both speakers A and B, the model cannot determine the speaker for whom it should provide a recommendation. Therefore, we incorporated speaker embedding \cite{sun2019dream} in the calculation of RoBERTa’s token embeddings.

In other words, given a token in the dialogue, the corresponding embedding is calculated using the formula $\bm{we}(x_i) + \bm{po}(i) + \bm{se}(Speaker(x_i))$, where $\bm{we}(\cdot)$ denotes word embedding, $\bm{po}(\cdot)$ is position embedding, and $\bm{se}(\cdot)$ is speaker embedding.
In addition, $Speaker(\cdot)$ is a function that returns $1$ if the input token is from the utterance of the speaker to be recommended and $0$ otherwise.

\paragraph{ChatGPT}
In ChatGPT, we substituted dialogue $U$ and tourist-spot description $e_i$ for a speaker summary and recommendation information in our baseline.
ChatGPT generated the score based on the assigned task definition sentence, a few-shot sample, a speaker summary denoted by $s_k(k=A,B)$, and a tourist-spot description $e_i$.
Table \ref{baseline_prompt} (Appendix \ref{Promt Design}) presents an example of the prompt.
We used the gpt-3.5-turbo-16k-0613 model with the temperature parameter set to 0.0.

\paragraph{Human}
We also compared the results with human predictions.
We used the estimated scores obtained from the five crowdworkers mentioned in section \ref{score}.

\begin{table*}[t]
\centering
\begin{tabular}{lllllllll}
\hline
\textbf{Topic} & \textbf{Method} & \textbf{N@1} & \textbf{N@3} & \textbf{N@5} & \textbf{R@1} & \textbf{R@3} & \textbf{R@5} & \textbf{Coef.}\\
\hline
ALL & Random & 0.497 & 0.531 & 0.569 & 0.059 & 0.191 & 0.318 & -0.025 \\
& RoBERTa & 0.605 & 0.621 & 0.648 & 0.083 & 0.230 & 0.374 & 0.156 \\
& SumRec (RoBERTa) & \textbf{0.615} & \textbf{0.626} & \textbf{0.655} & \textbf{0.092} & \textbf{0.241} & \textbf{0.376} & \textbf{0.173} \\
& ChatGPT & - & - & - & 0.063 & 0.196 & 0.317 & 0.077 \\
& SumRec (ChatGPT) & - & - & - & 0.074 & 0.207 & 0.335 & 0.112 \\
\hdashline
& Human & 0.612 & 0.640 & 0.656 & 0.089 & 0.245 & 0.374 & 0.183 \\
\hline
Travel & Random & 0.495 & 0.533 & 0.571 & 0.049 & 0.170 & 0.298 & -0.006 \\
& RoBERTa & 0.576 & 0.596 & 0.629 & 0.073 & 0.206 & \textbf{0.358} & \textbf{0.155} \\
& SumRec (RoBERTa) & \textbf{0.597} & \textbf{0.605} & \textbf{0.639} & \textbf{0.078} & \textbf{0.226} & \textbf{0.358} & 0.152 \\
& ChatGPT & - & - & - & 0.073 & 0.204 & 0.306 & 0.080 \\
& SumRec (ChatGPT) & - & - & - & 0.074 & 0.210 & 0.330 & 0.116 \\
\hdashline
& Human & 0.576 & 0.613 & 0.641 & 0.085 & 0.257 & 0.387 & 0.196 \\
\hline
Except for Travel & Random & 0.490 & 0.527 & 0.558 & 0.047 & 0.172 & 0.302 & -0.023 \\
& RoBERTa & 0.572 & 0.596 & 0.618 & 0.071 & 0.227 & 0.371 & 0.117 \\
& SumRec (RoBERTa) & \textbf{0.608} & \textbf{0.618} & \textbf{0.637} & \textbf{0.098} & \textbf{0.248} & \textbf{0.382} & \textbf{0.163} \\
& ChatGPT & - & - & - & 0.065 & 0.218 & 0.349 & 0.064 \\
& SumRec (ChatGPT) & - & - & - & 0.073 & 0.233 & 0.349 & 0.105 \\
\hdashline
& Human & 0.606 & 0.648 & 0.653 & 0.093 & 0.245 & 0.377 & 0.188 \\
\hline
No Restriction & Random & 0.496 & 0.539 & 0.576 & 0.051 & 0.173 & 0.301 & -0.009 \\
& RoBERTa & 0.632 & 0.646 & 0.669 & 0.087 & 0.239 & 0.375 & 0.172 \\
& SumRec (RoBERTa) & \textbf{0.651} & \textbf{0.661} & \textbf{0.678} & \textbf{0.096} & \textbf{0.251} & \textbf{0.385} & \textbf{0.183} \\
& ChatGPT & - & - & - & 0.058 & 0.186 & 0.317 & 0.081 \\
& SumRec (ChatGPT) & - & - & - & 0.068 & 0.196 & 0.333 & 0.117 \\
\hdashline
& Human & 0.634 & 0.652 & 0.667 & 0.088	& 0.239 & 0.367 & 0.176 \\

\hline
\end{tabular}
\caption{\label{automatic eval}
Automatic evaluation results for the ChatRec test set, where N@k, R@k, and Coef represent NDCG@k, Recall@k, and Spearman’s rank correlation coefficient, respectively.
}
\end{table*}

\subsection{Evaluation Metrics}
We used Normalized Discounted Cumulative Gain (NDCG@k) \cite{10.1145/582415.582418}), Recall@k, and Spearman’s rank correlation coefficient for the evaluation metrics.

NDCG, an evaluation metric for ranking recommendation performance and information retrieval, takes a value from 0 to 1, with a value closer to 1 indicating a correct ranking.
The calculation of NDCG@k is as follows :
\begin{equation}
NDCG@k = \frac{DCG_k}{IDCG_k}
\end{equation}
\begin{equation}
DCG_k = \sum^K_{i=1}\frac{2^{rel_i}-1}{log_2(i+1)}
\end{equation}
Here, $k$ is the number of top-ranked objects used for the $NDCG$ calculation.
In experiments, the values of $k$ are 1, 3, and 5; $rel_i$ is the evaluation score of the tourist spots at rank $i$.
The ideal $DCG$ ($IDCG$) is the $DCG$ if the ranking list has been correctly ordered by ground-truth scores.
The NDCG cannot assess tied rankings. Because ChatGPT yields a discrete value, it often predicts multiple tourist spots to hold identical ranks.
For the above reasons, NDCG cannot accurately evaluate ChatGPT’s performance.
Therefore, we did not assess ChatGPT based on NDCG.

Recall@k is often used as an evaluation metric in natural-language processing tasks.
It calculates the proportion of top-k items correctly ranked by a model or algorithm. 
This study categorized those with evaluation scores from 1 to 3 as ``incorrect” and those with scores from 4 to 5 as ``correct” within Recall@k.
We used 1, 3, and 5 for the $k$ values, which are consistent with the values in our NDCG@k evaluation.

Spearman’s rank correlation coefficient measures the relation between a model’s predicted ranking order and the actual order of correct responses.

\subsection{Results}
In addition to an overall analysis labeled ALL, which represents the average of all data, targeted assessments were conducted for each of the three topics—Travel, Except for Travel, and No Restriction.
Table \ref{automatic eval} presents the results of the automatic evaluation.

Findings show that SumRec (RoBERTa) with the proposed framework consistently outperformed the baselines in all metrics.
Similarly, a notable improvement in performance is observed for SumRec (ChatGPT) compared to the ChatGPT model, which did not employ the proposed framework.
Furthermore, SumRec (RoBERTa) outperforms human predictions on several metrics.
The speaker summaries likely increase attention toward the speaker’s tastes, preferences, and habits.
Similarly, recommendation information may facilitate the connection between speaker information and tourist-spot characteristics.
These results indicate that the SumRec framework effectively recommended tourist spots through chat dialogues.

Meanwhile, the performance enhancement seen in SumRec for Travel is not as notable as that for Except for Travel, which is likely attributed to the nature of the travel data. 
Because it comprises dialogues about travel, the model manages to capture some interaction between the speaker’s information and tourist-spot descriptions.
As such, adding recommendation information may not substantially increase the model’s performance.

Appendix \ref{Case Study} presents an example of the complete input and prediction scores for each method.

\section{Discussion}

\subsection{Quality of Speaker Summary and Recommendation Information}\label{human-eval}

We conduct a crowdsourcing evaluation to verify the quality of the speaker summary and recommendation information generated by ChatGPT.
Speaker summaries are assessed by focusing on two main aspects: whether they are consistent with the dialogue (Consistency) and whether they cover all essential information provided within that dialogue (Comprehensiveness).
Similarly, the evaluation of recommendation information emphasizes two aspects: its alignment with the described tourist spot (also referred to as Consistency) and the sufficiency of the provided details (Informativeness).
The workers rate each metric using a five-point Likert scale.
We choose 100 speaker summaries from the test data, ensuring as balanced a representation as possible of each test-set topic.
For recommendation information, we select 100 random samples from all tourist spots in the test data.
Four evaluators assess each speaker summary and the recommendation information; their average score serves as the final rating for each sample.

\begin{table}[t]
\centering
\begin{tabular}{l|lc}
\hline
\textbf{} & \textbf{Metric} & \textbf{Score} \\
\hline
Speaker  & Consistency & 4.24 \\
Summary& Comprehensiveness & 3.78 \\ \hline
Recommendation & Consistency & 4.70 \\
Information& Informativeness & 3.43 \\
\hline
\end{tabular}
\caption{\label{tab:human-eval}
Human evaluation results for speaker summaries and recommendation information generated by ChatGPT.
}
\end{table}

\begin{table}[t]
\centering
\begin{tabular}{p{0.97\linewidth}}
\hline
\textbf{Tourist Spot:} Shika-no-yu \\ 
\hline
\textbf{Description:} \\ 
\quad A day-trip bathing facility using the famous nigori-yu hot spring water from Shika-no-yu. This hot water is poured into the baths, and guests can rest in a private room for 2,000 yen per person for a minimum of two people for five hours from 10:00 to 17:00. Guests can use the open-air bath during business hours. \\ 
\hline
\textbf{Generated Recommendation Information:} \\ 
\quad This tourist spot is recommended for those who want to enjoy hot springs and day-trip bathing. This is especially suitable for those who want to enjoy Shika-no-yu’s famous hot spring water. Private rooms are also available for those who want to relax and unwind. There is also an open-air bath for guests who want to enjoy hot spring baths in the midst of nature. \\ 
\hline
\end{tabular}
\caption{
An example of a tourist spot description and the generated recommendation information.
}
\label{desc_rec}
\end{table}

\begin{table*}
\centering
\begin{tabular}{llllllll}
\hline
\textbf{Method} & \textbf{N@1} & \textbf{N@3} & \textbf{N@5} & \textbf{R@1} & \textbf{R@3} & \textbf{R@5} & \textbf{Coef.} \\
\hline
SumRec (RoBERTa) & \textbf{0.615} & \textbf{0.626} & \textbf{0.655} & \textbf{0.092} & \textbf{0.241} & \textbf{0.376} & \textbf{0.173} \\
w/o Sum. & 0.584 & 0.608 & 0.641 & 0.082 & 0.235 & 0.370 & 0.153 \\
w/o Rec. & 0.577 & 0.606 & 0.635 & 0.076 & 0.226 & 0.365 & 0.125 \\
5 turns & 0.600 & 0.622 & 0.646 & 0.085 & 0.239 & 0.375 & 0.158 \\
\hline
\end{tabular}
\caption{\label{turn study}
Comparison results for the ablation model. N@k, R@k, and Coef represent NDCG@k, Recall@k, and Spearman’s correlation coefficient, respectively.
}
\end{table*}

Table \ref{tab:human-eval} presents the results of human evaluation.
Although the speaker summaries and recommendations demonstrate a noticeable degree of coherence, the summaries’ comprehensiveness and the recommendations’ richness of information are slightly deficient. Several description sentences for tourist spots are notably brief.
Additionally, a substantial understanding of travel specifics is required to generate relevant recommendation information.
Table \ref{desc_rec} presents an example of tourist-spot description and recommendation information.
The description provides details about the facility as a day spa as well as the water temperature, operational hours, and cost.
However, generating recommendation information using such information as operational hours and cost yields limited explanatory value.

Results show that although the speaker summaries and recommendation information are highly consistent, the former’s comprehensiveness and the latter’s informativeness leave room for improvement.
This suggests that ChatGPT’s pretrained knowledge cannot solely generate informative recommendation information.
Future work must include the use of external knowledge to generate more relevant information.

\subsection{Ablation Study}
To better understand the effectiveness of each information in SumRec, which has demonstrated the best performance among all the tested methods, we conduct an ablation study by considering the following SumRec variants (RoBERTa).

\begin{itemize}
      \item \textbf{w/o Sum.}: This model does not use speaker summary. Instead, we directly input the dialogue $U$. The procedure for inputting dialogues into the score estimator mirrors the one used by RoBERTa in our comparative approach.
      \item \textbf{w/o Rec.}: In this variant, we remove recommendation infomation $r_i$.
      \item \textbf{5 turns}: This model uses the initial 5 turns (10 utterances) in dialogue to generate a speaker summary, which represents approximately half of the total dialogue. Summaries produced using this method might not be as comprehensive as those generated from complete dialogues.
\end{itemize}

Table \ref{turn study} presents the results for each ablation model.
SumRec (RoBERTa) delivered the highest performance for all the evaluation measures, indicating that both speaker summaries and recommendation information considerably enhance its performance.
A comparison between ``w/o Sum'' and ``w/o Rec'' reveals that the former performed better. 
Even without a speaker summary, acquiring information about the speaker from the dialogue is still possible.
Therefore, the effect of summaries seems relatively low, suggesting that recommendation information has greater importance.

The results for ``5 turns'' exhibited less decline in the performance compared to those for ``w/o Sum'' and ``w/o Rec.''
Summaries produced from dialogues based on only five turns may be less extensive and be of lower quality compared to those generated using all available turns.
Nonetheless, using a speaker summary prove to be more effective as ``5 turns'' outperformed the variant that lacked a summary.
This demonstrates the benefit of having a speaker summary regardless of its quality.

\section{Conclusion}
We defined a novel task where tourist spots are recommended based on chat dialogues.
Additionally, we proposed SumRec, a framework designed specifically for this task.

Moreover, we collected a dataset comprising human dialogues, tourist-spot information, and tourist-spot evaluation scores.
Our proposed framework summarized speaker characteristics from chat dialogues and generated recommendation information, which explains for whom tourist spots are recommended based on their descriptions.
The framework then employed this information to provide recommendations.
Experimental results indicated that our framework exhibits excellent performance on all the considered evaluation metrics.

Nevertheless, travel-related dialogues continue to show considerable potential for improvement.
We plan to explore more efficient methods by incorporating prior knowledge about the trip and integrating external knowledge into the model.

\section*{Acknowledgements}
This work was supported by JSPS KAKENHI Grant Number 19H05692.

\bibliography{anthology,custom}
\bibliographystyle{acl_natbib}

\appendix

\section{Prompt Examples}
\label{Promt Design}
\paragraph{Speaker Summary}
We design a one-shot prompt to allow ChatGPT to generate a speaker summary that extracts the speaker’s interests, preferences, and habits.
For generating speaker summaries, ChatGPT produces $s_A$ and $s_B$ based on a task description, an example of a dialogue, and its speaker summary as well as the target dialogue $U$.
The author prepares the one-shot speaker summary.
Table \ref{summary_prompt} presents an example of a speaker summary prompt.

\paragraph{Recommendation Information}

For recommendation information, we select a five-shot prompt because suitable sentences cannot be produced using fewer than four shots.
ChatGPT generates $r_i$ from a task description, five tourist-spot descriptions, examples of their recommendation information, and a description $s_i$ as the target.
The author also prepares five-shot prompts for recommendation information.
Table \ref{recommend_prompt} presents an example of a recommendation information prompt.

\paragraph{Score Estimator}

For the SumRec score estimator, we design a five-shot prompt for both the baseline ChatGPT and SumRec (ChatGPT).
The former generates scores based on the assigned task definition sentence, a few-shot sample, dialogues $U$, and a tourist-spot description $e_i$.
The latter (ChatGPT) receives the task definition sentence, a five-shot sample, speaker summary $s_k(k=A,B)$, and tourist-spot description $e_i$ as well as recommendation information $r_i$ and generates an estimated score.
We also create prompts by randomly selecting five examples for each topic to be fair to the model using RoBERTa.

Tables \ref{sumrec_prompt} and \ref{baseline_prompt} show examples of prompts for SumRec (ChatGPT) and the baseline ChatGPT, respectively.

\begin{table*}[t]
\centering
\small
\begin{tabular}{p{0.9\linewidth}}
\hline
\textbf{Prompt} \\
\hline
[Task] \\
Summarize the history of dialogue between Mr. A and Mr. B regarding their respective interests, habits, and profiles. \\
\\
--1-- \\ \relax
[Dialogue] \\
A:Quickly, do you have any hobbies? \\
B: I am an indoor person, so I play games all the time at home. What kind of hobbies do you have? \\
\dots \\
A:It is true that mountain climbing can be dangerous. I was once worried that I might have lost my way when I was climbing a mountain by myself, but I was really scared. \\
B:So this is your own experience. Scary. But I would like to climb a mountain by myself someday. \\
\\ \relax
[Summary] \\
A: \\
She is an indoor person. Her hobbies are things she can do at home, \dots \\
B: \\
He is an indoor person and enjoys playing games. \dots \\
\\
--2-- \\ \relax
[Dialogue] \\
A:Nice to meet you. \\
B:Nice to meet you. It's nice to meet you too. \\
A:Do you have any hobbies? \\
B:Well, I am not very active, but I do handicrafts and read books when I am alone. \\
A:Me too. I make handmade accessories. \\
B:Accessories? That's wonderful. What kind of items do you make exactly? \\
A:I often make earrings with beads. \\
B:Beads? I tried that once, but I couldn't get it right and gave up. \\
A:I see, what kind of items do you make? \\
B:Right now I am doing crocheting and stitching. The crochet is for cushion covers and the stitching is for kitchen cloths. \\
A:That's wonderful. It's nice to be able to make things you can use in your daily life. \\
B:Yes, you can never have too many dishcloths, so it's good that you don't waste time making them. \\
A:Maybe I should try it too! \\
B:Please give it a try. You can make all kinds of things with beads. Charms, earrings, and so on. \\
A:Yes, that's right. Rings are also fun and easy to make. \\
B:Rings are wonderful. Beads are also fun to color-match. I sometimes use them when I do embroidery. \\
A:Handicrafts are great because they allow me to be mindless. Bead embroidery is so wonderful! \\
B:Yes, it is. I feel like I can reset my mind when I do needlework.
A:There are times when I get frustrated because I can't get it right, but the joy I feel when I finish it makes it all go away. \\
B:Yes, even if I get frustrated while working on it, in the end I feel a sense of accomplishment and a sense of having done it. \\
A:It's nice to talk to a nice person. \\
B:It was great fun talking about handicrafts with you too. Thank you very much. \\
\\ \relax
[Summary] \\
A: \\
\hl{She enjoys handicrafts and handmade goods, and is especially good at creating accessories. She often makes earrings using beads. She says she enjoys making things that are handmade and usable in everyday life.} \\
\hl{B: \\
She enjoys doing handicrafts and reading in her spare time when she is alone. Recently, she has been crocheting and stitching, making cushion covers with her crochet and dishcloths with her stitching. She says she feels like she can reset her mind by doing handicrafts.}  \\ \\
\hline
\end{tabular}
\caption{
Prompt sample for summarizing speaker information. Highlights show the passages generated via ChatGPT
}
\label{summary_prompt}
\end{table*}

\begin{table*}[t]
\centering
\small
\begin{tabular}{p{0.9\linewidth}}
\hline
\textbf{Prompt} \\
\hline
[Task] \\
From the description of the tourist spot, please explain whom it is recommended for. \\
 \\
--1-- \\ \relax
[Descriptions of tourist spots] \\
Aqua City Odaiba \\
A large shopping complex overlooking the center of the city over the Statue of Liberty   \dots   The upper floor also houses the Aqua City Odaiba Shrine, which is very popular with families. \\
 \\ \relax
[Recommendation Information] \\
This tourist destination is recommended for those who want to enjoy shopping and dining.  \dots \\
 \\
\dots \\
\\
--5-- \\ \relax
[Tourist Spot Descriptions] \\
Cape Notorimisaki \\
Located on the east side of Lake Nodori, Cape Nodori is famous for its sunset and starlit sky. The area consists of cliffs 40–50m high and is famous as a filming location for movies and commercials. In the winter, drift ice comes in contact with the shore.
 \dots \\
 \\ \relax
[Recommendation Information] \\
This tourist spot is ideal for those who want to enjoy beautiful sunsets and starry skies,
This place is recommended for those who want to enjoy beautiful sunsets and starry skies. It is also famous as a filming location for movies and commercials, so those who enjoy photography and filming can also enjoy this place. It is also recommended for those who want to enjoy the winter scenery as drift ice comes in contact with the shore in winter.  \\
 \\
--6-- \\ \relax
[Tourist Spot Descriptions] \\
Kobe Animal Kingdom \\
Located on Kobe's Port Island, Kobe Animal Kingdom is a park where visitors can enjoy interacting with different animals, including alpacas and capybaras, amid flowers in full bloom all year round. The powerful and moving bird performance show is very popular. The park is all-weather, even on rainy days. The park is also barrier-free, with a flat terrain and restrooms for the physically challenged. It is a spot that can be enjoyed by all family members and friends. \\
 \\ \relax
[Recommendation Information] \\
\hl{Recommended for people who want to interact with animals and flowers.It is especially enjoyable for families and friends. The facility is also barrier-free, so it is a friendly environment for people with disabilities. The bird performance show is very popular, so it is also recommended for people who like entertainment. }\\
\hline
\end{tabular}
\caption{\label{recommend_prompt}
Prompt sample for generating recommendation information. Highlights denote generated text by ChatGPT.
}
\end{table*}

\begin{table*}
\centering
\small
\begin{tabular}{p{0.9\linewidth}}
\hline
\textbf{Prompt} \\
\hline
\# Task \\
Based on the characteristics of the speaker, the description of the tourist spot, and the description of the recommended person, choose one of the following preferences for the tourist spot. \\
1: I definitely don't want to go \\
2: I don't want to go \\
3: Neither \\
4: I want to go \\
5: I definitely want to go \\
 \\
\#Begin Assessment \\
--1-- \\ \relax
[Characteristics of Speakers] \\
She mentions that spring is coming and talks about the cold.  \dots  but she talks about her everyday life with her daughter as if it were a real game. \\
 \\ \relax
[Tourist Spot Descriptions] \\
The museum is surrounded by beautiful gardens with a view of Bandaisan.  \dots  The required time is 1 hour and 30 minutes.  \\
 \\ \relax
[Recommended Person] \\
This tourist destination is recommended for those who are interested in fine arts and culture.  \dots  It is recommended for those who want to take their time to appreciate fine arts and those who want to take their time to appreciate the works. \\
 \\ \relax
[Score] \\
5 \\
 \\
 \dots \\
  \\
--5-- \\ \relax
[Characteristics of Speakers] \\
She tells me that in the cold season she craves to eat nabe.  \dots  Can. She likes watching sports, especially professional baseball and sumo. \\
 \\ \relax
[Tourist Spot Descriptions] \\
A complex facility that consists of a restaurant zone where you can enjoy seafood such as pufferfish,  \dots  it is a popular spot where you can enjoy Shimonoseki specialties. \\
 \\ \relax
[Recommended Person] \\
This tourist spot is recommended for those who want to enjoy seafood and local specialties.  \dots  You can also enjoy the beautiful scenery of the Kanmon Strait, so it is also suitable for those who want to taste local specialties while enjoying the scenery. \\
 \\ \relax
[Score] \\
1 \\
 \\
--6-- \\ \relax
[Characteristics of Speakers] \\
I like reptiles and frogs, so I collect and keep them. I especially like horned frogs, and I take care of several kinds of them. Many horned frogs live overseas, and I often buy them from breeders. They also have various patterns and colors, and the rare ones are especially expensive. Raising them takes a lot of time and effort in the winter, such as laying a panel heater and changing the water, but the cost is unexpectedly low. \\
 \\ \relax
[Tourist Spot Descriptions] \\
Upstream of the Ishikarigawa River, there is a columnar joint cliff that continues for 24km. Many falls flow down from the snow valley, and together with the surrounding primeval forest, it is an amazing landscape. It was once called Sounbetsu ( a river with many falls) in the Ainu language, so in 1921, it was named Soun-kyo by the critic - essayist Omachi Keigetsu. \\
 \\ \relax
[Recommended Person] \\
This tourist spot is recommended for those who want to enjoy the beauty of nature and the scenery of the falls. Specifically, it is the best place for those who want to enjoy the scenery of the cliffs and primeval forests. It is also recommended for those who love photography and those who like nature observation. You can also enjoy outdoor activities and hiking, so it is suitable for those who like nature and an active way of life. \\
 \\ \relax
[Score] \\
\hl{5} \\
\hline
\end{tabular}
\caption{
Prompt sample for SumRec. Highlights denote generated score by ChatGPT.
}
\label{sumrec_prompt}
\end{table*}

\begin{table*}
\centering
\small
\begin{tabular}{p{0.9\linewidth}}
\hline
\textbf{Prompt} \\
\hline
\# Task \\
Based on the conversation history and the description of the tourist spot, please select from the following five choices how much Mr. A and Mr. B like the tourist spot. \\
1: I definitely don't want to go \\
2: I don't want to go \\
3: Neither \\
4: I want to go \\
5: I definitely want to go \\
 \\
\# Begin Assessment \\
--1-- \\ \relax
[Dialogue] \\
A: Hello. Do you have any holiday plans? \\
B: Hello. I think I will go to Sotobo in Chiba on holiday this week. \\
A: I see. By the way, have you ever attended your own coming-of-age ceremony? \\
B: No, it was a long time ago, but I was sick on the day and was absent. How was it? \\

\dots \\

A: That's right. But I don't usually sing Western music because I don't get excited when I sing it, which nobody knows. \\
B: That may be true. \\
A: Do you order meals other than drinks at karaoke? \\
B: I used to order rice and noodles, but these days I only order snacks. \\ \relax
[Tourist Spot Descriptions] \\
The entire park is the culmination of the basic design of world-renowned sculptor Isamu Noguchi. At 1.89 million square meters,  \dots there is also a restaurant where you can enjoy French cuisine made with fresh ingredients from Hokkaido. \\ \relax
[Score] \\
A: 2 \\

\dots \\
 \\
--5-- \\ \relax
[Dialogue] \\
A: It's really cold. It's snowing here right now. \\
B: I see. It seems to have rained down here only once at the end of the year, but I haven't seen it. Is it piled up? \\
\dots \\
A: Yes. Thank you for the good time. \\ \\
B: Thank you very much. \\ \relax
[Tourist Spot Descriptions] \\
It is located in BIG FUN Heiwa Island where there are movie theaters, - Game Center, -  \dots  and there is a package plan that includes picking up and dropping off from the airport and bathing along with the flights arriving late at night and departing early in the morning. \\ \relax
[Score] \\
A: 5 \\
 \\
--6-- \\ \relax
[Dialogue] \\
A: Where would you like to go? \\
B: I like Okinawa. The sea is very beautiful. \\
A: Are you more of a sea person than a mountain person? I went to a remote island in Okinawa only once, and I am very impressed. \\
B: You're from the sea. I like the remote islands. By the way, where did you go? \\
A: I went around Ishigaki Island and three other islands. \\
\dots \\
A: I'm curious to go to a country I've never been to. But I miss Europe because of COVID-19, so I might go to France first. \\
B: Yes. I hope that the COVID-19 will be brought under control soon. \\
 \\ \relax
[Tourist Spot Descriptions] \\
It is located in the center of Shiba Park, and is the head temple of the Jodo sect established in 1393. In the Edo era,  \dots The Sangedatsumon (important cultural property) that was built in 1622 is a building where you can remember the great view of the past. \\
 \\ \relax
[Score] \\
A: \hl{2} \\
\hline
\end{tabular}
\caption{
Prompt sample for the baseline. Highlights denote generated score via ChatGPT.
}
\label{baseline_prompt}
\end{table*}

\section{Case Study} \label{Case Study}
Table \ref{case_study} presents the dialogues, tourist-spot description, speaker summary and recommendation information, and scores estimated using various methods.
SumRec easily combines speaker information with the tourist-spot features using the speaker summary and recommendation information and successfully estimates high and low evaluation scores by each speaker.
For tourist spots 1 and 2, most methods estimate a higher score for Speaker A than for Speaker B based on the knowledge that Speaker A is interested in historical buildings. 
For tourist spot 3, Speaker A did not specifically mention this type of tourist spot, while Speaker B stated that they do not like traveling in large groups.
Based on this information, SumRec (RoBERTa) succeeds in estimating a lower score for Speaker B than for Speaker A. Meanwhile, even ChatGPT, which employs SumRec, could not account for this difference.
For humans, the estimated scores for Speakers A and B show the exact opposite relation.

\begin{table*}
\centering
\tiny
\begin{tabularx}{\linewidth}{p{0.0728\linewidth}p{0.08\linewidth}p{0.768\linewidth}}
\hline
Speaker Info. & Dialogue & A:Nice to meet you. \\
& & B:Nice to meet you too. \\
& & A:It's getting warmer, isn't it? \\
& & B:Yes, it is. It's hard to go out in COVID-19, but have you been out lately? \\
& & A:I haven't been out much lately. I would like to go for a drive somewhere. \\
& & B:A drive sounds nice. Here in COVID-19, the lawn cherry blossoms and tulips are in season, so I go out often. \\
& & A:That sounds nice. I love nature. Do you like to be with friends? Or do you prefer to stay in peace and quiet? \\
& & B:I like to stay in and relax. I don't like to move with a lot of people. Did you go cherry blossom viewing? \\
& & A:I see. I didn't go to Hanami this year. \\
& & B:I see. Did you spend more time at home in COVID-19? \\
& & A:Yes, it has increased. I am hesitant to go out. How about you? \\
& & B:That's right. I can't go to places where there are many people, so I tend to have too much time on my hands. \\
& & A:I see. By the way, do you like historical buildings? Or do you like new things? \\
& & B:I like new things more, like theme parks using VR and the latest technology! I'm excited by things like theme parks using VR and the latest technology. How about you? \\
& & A:I like castles, shrines, and old buildings. I think VR technology will become mainstream in the future. I am studying it. \\
& & B:I see. If you could go out freely now, is there a place you would like to visit? Is there a place you would like to go if you could go out freely? \\
& & A:I like to be in touch with nature, so any place like that! Nowadays, you can even travel in VR, can't you? \\
& & B:Yes, that's true. It's not quite the same as going there, but watching VR sightseeing videos is quite relaxing. \\
& & A:It's nice, isn't it? I never get tired of watching videos shot with a drone. It makes me want to stay at home even more, though. \\
& & B:So that's one way to do it! Even if you can't go outside, there are things you can enjoy unexpectedly, aren't there? \\
& & A:We live in an era of convenience. Last but not least, are you a mountain person? Or do you prefer the ocean? I love both! \\
& & B:I like plants and so on, so I am definitely a mountain person! It's great that you can enjoy both. \\
& & A:I see. I hope you will be able to go out freely soon! Thank you very much for your time today. \\
& & B: That's so true. Thank you very much for your time as well. \\
\hdashline
& Speaker Summary & A:\quad \textcolor{blue}{I love nature and enjoy driving and going to the mountains and the sea}. \textcolor{blue}{I am also interested in historical and old buildings}. Recently, I have been hesitant to go out because of COVID-19, so I use VR and drones to feel like I am traveling. \\
&  & B:\quad I like nature and often go out when the lawn cherry blossoms and tulips are at their best. \textcolor{red}{I am not good at moving in large groups}, so I like to stay in and relax. I am also interested in new technology and am excited about VR and the latest theme parks. Since my outings are restricted due to COVID-19, I use VR and drones to get a travel feel. \\
\hline
Tourist Spot 1 & Name & Mt. Ontake \\
& Description & Ontake is 929m above sea level, and with a cable car and well-maintained trails, hikers can easily enjoy hiking. Musashi Ontake Shrine stands at the summit. Just below the summit is the village of Ontake, which is home to many inns and eateries that serve as guides for religious climbers. \\
\hdashline
& Recommendation Information & This tourist destination is recommended for \textcolor{blue}{those who want to enjoy nature and hiking}. It is conveniently accessible by cable car, making it easy to enjoy climbing the mountain. The Musashi Ontake Shrine is also located at the summit, making it ideal for those who want to enjoy religious mountain climbing. Furthermore, there are many lodgings and eating establishments clustered just below the summit, making it a great place for those who want to spend a relaxing time. \\
\hdashline
& Speaker & A:\quad5 \\
& Score & B:\quad2 \\
\hdashline
& SunRec & A:\quad3.89 \\
& (RoBERTa) & B:\quad3.6 \\
\hdashline
& SunRec & A:\quad4.0 \\
& (ChatGPT) & B:\quad4.0 \\
\hdashline
& Human & A:\quad4.0 \\
&  & B:\quad3.8 \\
\hline
Tourist Spot 2 & Name & Sengakuji Temple \\
& Description & The Sengakuji Temple is a Soto sect temple famous for the graves of the 47 ronin. The main hall also houses the guardian deity of Oishi Kuranosuke. In 2001, the Ako Yoshishi Memorial Hall was built to commemorate the 300th anniversary of the Ako raid, and contains items left behind by the warriors. The Yoshishi Festival is held twice a year in spring and winter, and an exhibition of temple treasures is held during the spring festival. \\
\hdashline
& Recommendation Information & This tourist spot is recommended for those \textcolor{blue}{interested in Japanese history and culture}. In particular, Sengakuji Temple is recommended for those interested in the 47 ronin and Ako Yoshishi. The temple also holds events such as exhibitions of temple treasures and the Yoshishi Festival, giving visitors the opportunity to experience history. The temple is also known as a temple of the Soto sect of Buddhism and is suitable for those interested in Buddhism. \\
\hdashline
& Speaker & A:\quad5 \\
& Score & B:\quad1 \\
\hdashline
& SunRec & A:\quad3.62 \\
& (RoBERTa) & B:\quad3.61 \\
\hdashline
& SunRec & A:\quad4.0 \\
& (ChatGPT) & B:\quad3.0 \\
\hdashline
& Human & A:\quad3.8 \\
&  & B:\quad 2.6 \\
\hline
Tourist Spot 3 & Name & Tokyo Dome \\
& Description & The core of Tokyo Dome City. Japan's first all-weather multipurpose stadium. Known not only for baseball but also as a venue for \textcolor{red}{various events}. \\
\hdashline
& Recommendation Information & This tourist spot is recommended for people who like to watch sports and events. It is a must-see for baseball fans, and is also a great place for those who want to enjoy entertainment as it hosts a variety of events. It is also a great place for music lovers, as concerts and live music events are also held here. \\
\hdashline
& Speaker & A:\quad3 \\
& Score & B:\quad1 \\
\hdashline
& SunRec & A:\quad3.05 \\
& (RoBERTa) & B:\quad3.04 \\
\hdashline
& SunRec & A:\quad3.0 \\
& (ChatGPT) & B:\quad3.0 \\
\hdashline
& Human & A:\quad1.4 \\
&  & B:\quad 2.2 \\
\hline
\end{tabularx}
\caption{\label{case_study}
An example of ChatRec and estimated scores through various methods.
}
\end{table*}

\end{document}